\documentclass[journal]{IEEEtran}

\usepackage{mathtools}
\usepackage[makeroom]{cancel}
\usepackage{amssymb}
\usepackage{graphicx}
\usepackage[english]{babel}
\usepackage{caption}
\usepackage{thmtools}
\usepackage{color}
\usepackage[font=footnotesize, margin=1cm]{caption}
\usepackage[framemethod=tikz]{mdframed}
\usepackage{lipsum}
\usepackage{comment}
\usepackage{bm}
\usepackage{tabularx}
\usepackage{lipsum}       
\usepackage{changepage}   
\usepackage{amsmath}
\usepackage[tight,footnotesize]{subfigure}
\usepackage{multirow}

\DeclareMathOperator*{\argmax}{argmax}
\DeclareMathOperator*{\diag}{diag}
\newcommand{\nn}{\nonumber}
\newcommand{\mc}{\mathcal}

\newcommand{\mbb}{\mathbb}

\newcommand{\beq}{\begin{equation}}
\newcommand{\eeq}{\end{equation}}

\IEEEoverridecommandlockouts

\usepackage{ifthen}
\newboolean{showcomments}
\setboolean{showcomments}{true}
\newcommand{\yue}[1]{\ifthenelse{\boolean{showcomments}}
{ \textcolor{red}{(Yue says:  #1)}}{}}

\newcommand{\BEAS}{\begin{eqnarray*}}
\newcommand{\EEAS}{\end{eqnarray*}}
\newcommand{\BEQ}{\begin{equation}}
\newcommand{\EEQ}{\end{equation}}
\newcommand{\BIT}{\begin{itemize}}
\newcommand{\EIT}{\end{itemize}}

\newcommand{\prob}{\mathbb{P}}
\newcommand{\expec}{\mathbb{E}}

\newtheorem{RK}{Remark}

\usepackage{psfrag}

\usepackage{nomencl}

\makenomenclature

\usepackage{makeidx}
\makeindex

\usepackage{tikz}
\usetikzlibrary{shapes,backgrounds,calc}

\makeatletter
\tikzset{circle split part fill/.style  args={#1,#2}{%
 alias=tmp@name, 
  postaction={%
    insert path={
     \pgfextra{%
     \pgfpointdiff{\pgfpointanchor{\pgf@node@name}{center}}%
                  {\pgfpointanchor{\pgf@node@name}{east}}%
     \pgfmathsetmacro\insiderad{\pgf@x}
      \fill[#1] (\pgf@node@name.base) ([xshift=-\pgflinewidth]\pgf@node@name.east) arc
                          (0:180:\insiderad-\pgflinewidth)--cycle;
      \fill[#2] (\pgf@node@name.base) ([xshift=\pgflinewidth]\pgf@node@name.west)  arc
                           (180:360:\insiderad-\pgflinewidth)--cycle;            
         }}}}}
 \makeatother

\usepackage{pgffor}

\def\PP{\prob}
\def\E{\expec}

\def\s{\sigma}

\def\y{\bm{y}}
\def\P{\bm{P}}
\def\Q{\bm{Q}}
\def\s{\bm{s}}
\def\h{\bm{h}}
\def\v{\bm{v}}
\def\V{\bm{V}}
\def\Y{\bm{Y}}
\def\be{{\bm{\beta}}}
\def\thet{{\bm{\theta}}}

\usepackage{algorithmic}
\usepackage{arydshln}

\renewcommand\baselinestretch{1} 

\begin{document}

\title{A Learning-to-Infer Method for Real-Time \\Power Grid Multi-Line Outage Identification} 

\author{Yue~Zhao,~\IEEEmembership{Member, IEEE,}
        Jianshu~Chen,~\IEEEmembership{Member, IEEE,}
        and~H.~Vincent~Poor,~\IEEEmembership{Fellow, IEEE}
\thanks{Some preliminary results from this work were presented in part at the IEEE Workshop on Statistical Signal Processing, Palma de Mallorca, Spain, 2016 \cite{SSP} and in part at the IEEE Global Conference on Signal and Information Processing (GlobalSIP), Arlington, VA, USA, 2016 \cite{zhao2016efficient}. This material is based in part on work supported by the SBU-BNL SEED Grant, and in part on work supported by the National Science Foundation under Grants DMS-1736417 and ECCS-1824710.}%
\thanks{Y. Zhao is with the Dept. of Electrical and Computer Engineering, Stony Brook University, Stony Brook, NY, 11794 USA (e-mail: yue.zhao.2@stonybrook.edu).}%
\thanks{J. Chen is with the Tencent AI Lab, Bellevue, WA, 98004 USA. (e-mail: jianshuchen@tencent.com)}%
\thanks{H. V. Poor is with the Dept. of Electrical Engineering, Princeton University, Princeton, NJ 08544 USA (e-mail:
poor@princeton.edu)}
}

\maketitle

\begin{abstract}
Identifying a potentially large 
number of 
simultaneous line outages in power transmission 
networks in real time is a computationally hard problem. This is because the number of hypotheses grows exponentially with the network size. A new ``Learning-to-Infer'' 
method is developed for efficient inference of every line status in the network. Optimizing the line outage detector 
is transformed to and solved as a discriminative learning problem based on Monte Carlo samples generated with power flow simulations. A major advantage of the developed Learning-to-Infer method is that the labeled data used for training can be generated in an arbitrarily large amount rapidly and at very little cost. As a result, the power of offline training is fully exploited to learn very complex classifiers for effective real-time multi-line outage identification. The proposed methods are evaluated in the IEEE 30, 118 and 300 bus systems. Excellent performance in identifying multi-line outages 
in real time is achieved with 
a reasonably small amount of data. 
\end{abstract}

\begin{IEEEkeywords}
Line outage detection, power system monitoring, 
machine learning, variational inference, Monte Carlo method 
\end{IEEEkeywords} 

\section{Introduction}
\label{sec:intro}

Lack of situational awareness in abnormal system conditions is a major cause of blackouts in power networks \cite{black04}. 
Network component failures such as transmission line outages, if not rapidly identified and contained, can quickly escalate to cascading failures. 
In particular, when line failures happen, the power network topology changes instantly, newly stressed areas can unexpectedly emerge, and subsequent failures may be triggered that lead to increasingly complex network topology changes. 
While the power system is usually protected against the so called ``$N-1$'' failure scenarios (i.e., only one component fails), as failures accumulate, effective automatic protection is no longer guaranteed. Thus, when cascading failures start developing, effective protective actions/interventions critically depend on correct and timely knowledge of the network status. 
Indeed, without accurate knowledge of the line outages, 
protective control methods have been observed to further aggravate the failure scenarios \cite{ferc2012arizona}. Thus, real-time line outage 
identification is essential to all network control decisions for mitigating failures. In particular, since the first few line outages may have already escaped the operators' attention, 
the ability to identify in real time the network topology with an \emph{arbitrary} number of line outages becomes critical to prevent system collapse. 

Real-time line outage 
identification is however a very challenging problem, especially when unknown line outages 
in the network quickly accumulate as in scenarios that cause large-scale blackouts \cite{black04}. \emph{The number of possible outage hypotheses 
grows exponentially with the number of line outages, 
making real-time multi-line outage 
identification fundamentally hard}. 
Other limitations in practice such as behaviors of human operators under time pressure, missing and contradicting information, and privacy concerns over data sharing can make this problem even harder. 
Assuming a small number of line failures, exhaustive search methods have been developed in \cite{lineoutage}, \cite{doubleline}, \cite{JSTSP14} and \cite{ZGP12} based on hypothesis testing, and in 
\cite{garcia2016line} based on logistic regression. 
To overcome the prohibitive computational complexity of exhaustive search methods, \cite{Hao} has developed sparsity exploiting outage identification methods with overcomplete observations 
to identify sparse multi-line outages. 
Without assuming sparsity of line outages, a graphical model based approach has been developed for identifying arbitrary multi-line outages 
\cite{chen2014line}. Sequential line outage detection method has also been proposed \cite{heydari2017quickest}. 

On a related note, \emph{non-real-time} power grid topology identification has also been extensively studied: the underlying topology stays the same, while many data are collected over a relatively long period of time before the topology can be identified \cite{li2013blind, yuan2016inverse, gera2017blind}. A variety types of data have been exploited for addressing this problem, e.g., data of power injections \cite{he2011dependency}, voltage correlation \cite{bolognani2013identification}, and energy prices \cite{kekatos2016online}. For power distribution systems in particular, various graph learning 
approaches have also been developed \cite{weng2017distributed, deka2017structure}. 

In this paper, we focus on \emph{real-time} identification of 
\emph{a potentially large number of} simultaneous line outages based on a set of measurements collected at any one point of time in the power system. We start with a probabilistic model of the variables in a power system (line statuses, power injections, voltages, power flows, currents etc.) and in its monitoring system (sensor measurements on all kinds of physical quantities). 
We then formulate the multi-line outage identification problem in a Bayesian inference framework, where we aim to compute the posterior probabilities of the post-outage topologies given any measurements at any one point of time. 

To overcome the fundamental computational complexity due to the \emph{exponentially large number of possible post-outage topologies}, 
we develop a learning based framework inspired by (but different from) variational inference, 
in which we aim to approximate the desired posterior probabilities using models that allow \emph{computationally easy marginal inference of line statuses}. 
Importantly, we develop ``end-to-end'' predictor models for multi-line outage identification, and allow arbitrary model structures and complexities. 
In order to find effective end-to-end predictor models, we transform optimizing a predictor model to a discriminative learning problem leveraging a Monte Carlo approach: 
a) based on \emph{full-blown power flow equations}, data samples of network topology, network states, and sensor measurements in the network can be efficiently generated according to a \emph{generative model} of these quantities, and b) with these simulated data, \emph{discriminative models} are learned \emph{offline}, which then offer \emph{real-time} prediction of the line outages 
based on newly observed 
measurements from the real network. We thus term the proposed method \emph{``Learning-to-Infer''}. It is important to note that this Learning-to-Infer method is not limited by any potential lack of real-world data, as the offline training procedure can be conducted entirely based on simulated data. 

A major strength of the proposed Learning-to-Infer method is that the \emph{labeled data set} for training the predictor model can be generated in an arbitrarily large amount, at very little cost. As such, we can fully exploit the benefit of offline model training in order to get accurate online multi-line outage 
identification performance. The proposed approach is also not restricted to specific models and learning methods, but can exploit any powerful models such as deep neural networks \cite{lecun2015deep}. As a result, predictor models of very high complexities can be adopted, yet without worrying about overfitting since more labeled training data can always be generated had overfitting been observed. 

The developed Learning-to-Infer method is evaluated in the IEEE 30, 118, and 300 bus systems \cite{UWEE} for identifying  
an \emph{arbitrary} number of line outages. 
It is demonstrated that, even with relatively simple predictor models and a reasonably small amount of data, the performance is surprisingly good for this very challenging task. 

The remainder of the paper is organized as follows. Section \ref{sec:prob} introduces the system model, and formulates real-time multi-line outage 
identification as a Bayesian inference problem. Section \ref{sec:lti} develops the Learning-to-Infer 
method. Section \ref{sec:NN} discusses the architectures of neural networks employed in this study. Section \ref{sec:num} presents the results from our numerical experiments. Section \ref{sec:concl} concludes the paper. 

\section{Problem Formulation}
\label{sec:prob}
\subsection{Power Flow Models}
We consider a power system with $N$ buses, and its \emph{baseline topology} (i.e., the network topology when there is no line outage) with $L$ lines. We denote the incidence matrix of the baseline topology by $M\in\{-1,0,1\}^{N\times L}$ \cite{ahuja1993network}. We use a binary variable $s_l$ to denote the status of a line $l$, with $s_l = 1$ for a connected line $l$, and $0$ otherwise. The actual topology of the network can then be represented by $\s = [s_1, \ldots, s_L]^T$. Generalizing this notation with a bit of abuse, we also employ $s_{mn}\in\{1,0\}$ to denote whether two buses $m$ and $n$ are connected by a line or not, (for simplicity, we consider any two buses can be connected by at most one line.) 
Given a network topology $\s$, the system's bus admittance matrix $\Y$ can be determined accordingly with the physical parameters of the system \cite{glover2011power}: $Y_{mn} = s_{mn}\left(G_{mn} + j B_{mn}\right)$, where $G_{mn}$ and $B_{mn}$ denote conductance and susceptance, respectively. Note that, when two buses $m$ and $n$ are not connected, $Y_{mn} = s_{mn} = 0$. 

We denote the real and reactive power injections at all the buses by $\P, \Q \in \mbb{R}^{N}$, and the voltage magnitudes and phase angles by $\V,\thet \in \mbb{R}^{N}$. Given the bus admittance matrix $\Y$, the nodal power injections and the nodal voltages satisfy the following AC power flow equations \cite{glover2011power}: $\forall m = 1,\ldots,N,$
\begin{align} \label{acpf}
P_m \!&=\! V_m \!\sum_{n=1}^N V_n s_{mn}\left(G_{mn}\!\cos(\theta_m-\theta_n) \!+\! B_{mn}\! \sin(\theta_m-\theta_n)\right), \nn\\
Q_m \!&=\! V_m \!\sum_{n=1}^N V_ns_{mn}\left(G_{mn}\!\sin(\theta_m-\theta_n) \!-\! B_{mn} \!\cos(\theta_m-\theta_n)\right), 
\end{align}
where a subscript $m$ denotes the $m^{th}$ component of a vector. 
In particular, given the network topology $\s$ and a set of controlled input values $\{\P, \Q^{in}, \V^{in}\}$, (where $\Q^{in}$ and  $\V^{in}$ consist of some subsets of $\Q$ and $\V$, respectively,) the remaining values of $\{\Q,\V,\thet\}$ can be determined by solving \eqref{acpf}. Typically, apart from a slack bus, most buses are ``$PQ$ buses'' at which the real and reactive power injections are controlled inputs, and the remaining buses are ``$PV$ buses'' at which the real power injection and voltage magnitude are controlled inputs \cite{glover2011power}. We refer the readers to \cite{glover2011power} for more details of solving AC power flow equations. 

A useful approximation of the AC power flow model is the DC power flow model: under a topology $\s$, the nodal real power injections and voltage phase angles approximately satisfy the following equation \cite{glover2011power}: 
\begin{align}
\P = MS\Gamma M^T \bm{\theta}, \label{DC}
\end{align}
where $S = \diag(s_1,\ldots,s_L)$, $\Gamma=\diag(\frac{1}{x_1}, \ldots, \frac{1}{x_L})$, and $x_l$ is the reactance of line $l$. We note that, in the DC power flow model, reactive power is not considered, and all voltage magnitudes are approximated by a constant. 

\subsection{Observation Models}
To monitor the power system, we consider real-time measurements taken by sensors measuring nodal voltage magnitudes and phase angles, current magnitudes and phase angles on lines, real and reactive power flows on lines, nodal real and reactive power injections, etc. In general, the observation model can be written as the following, 
\begin{align}
\y = \h(\s,\P,\Q^{in}, \V^{in})+ \v, \label{acobs}
\end{align}
where a) $\y\in\mbb{R}^K$ collects all the noisy measurements, b) $\h(\s,\P,\Q^{in}, \V^{in}) = \big[h_1(\s,\P,\Q^{in}, \V^{in}), \ldots$, $ h_K(\s,\P,\Q^{in}, \V^{in})\big]^T$ denotes the \emph{noiseless} values of the measured quantities, and the forms of $\{h_k(\cdot)\}$ depend on the specific locations and types of the sensors, and c) $\v$ denote the measurement noises. 

\begin{RK} A noiseless measurement function $h_k(\s,\P,\Q^{in}, \V^{in})$ can be an \emph{implicit function} 
without a closed form expression. For example, given $\s,\P,\Q^{in}$ and $\V^{in}$, while the  nodal voltage magnitude and phase angle at a particular $PQ$ bus can be solved from \eqref{acpf}, such a solution can only be obtained using numerical methods, and a closed form expression is not available. For discussions on the existence and uniqueness of the solution to the power flow equations \eqref{acpf}, we refer the readers to \cite{baldick2006applied}. 
\end{RK}

The observation models can be significantly simplified under the approximate DC power flow model \eqref{DC}. For example, measurements of $\bm{\theta}$ provided by phasor measurement units (PMUs) located at a subset of the buses $\mc{M}$ can be modeled as 
\begin{align}
\y = \bm{\theta}_\mc{M} + \bm{v}, \label{obs}
\end{align}
where $\bm{\theta}_\mc{M}$ is formed by entries of $\bm{\theta}$ from buses in $\mc{M}$. 
From the DC power flow model \eqref{DC}, we have 
\begin{align}
\thet = \left(MS\Gamma M^T \right)^+ \P, \label{thetaobs}
\end{align}
where $(\cdot)^+$ denotes pseudoinverse\footnote{For a connected network, the solution of $\thet$ given $\P$ is made unique by setting the phase angle at a reference bus to be zero.}. We note that, while the noiseless voltage phase angle measurements enjoy a closed form \eqref{thetaobs} and are linear in the power injections $\P$, they are \emph{not} linear in the line statuses $\s ~(=\diag(S))$. 

\subsection{Multi-Line Outage Identification as Bayesian Inference}
We are interested in identifying the post-outage network topology $\s$ \emph{in real time} based on instant measurements $\y$ collected in the power system. 
We formulate this multi-line outage identification problem as a \emph{Bayesian inference} problem. 
First, we model $\s, \bm{P}, \Q^{in}, \V^{in}$ and $\bm{y}$ with a joint probability distribution, 
\begin{align}
&p(\s, \bm{P}, \Q^{in}, \V^{in}, \bm{y}) \nn\\
&~~~~~~~~ = p(\s,\bm{P}, \Q^{in}, \V^{in}) \cdot p(\y|\s, \bm{P}, \Q^{in}, \V^{in}). \label{genmod}
\end{align} 
It is important to note that, given $\s, \bm{P}, \Q^{in}, \V^{in}$, the \emph{noiseless} measurements $\h$ (cf. \eqref{acobs}) can be \emph{exactly computed} by solving the AC power flow equations \eqref{acpf}. Adding noises to $\h$ then leads to $p(\y|\s, \bm{P}, \Q^{in}, \V^{in})$. 

\begin{RK}[Generative Model] \label{rkgen}
\eqref{genmod} represents a \emph{generative model} \cite{bishop2006pattern} with which a) the topology and the controlled inputs of power injections and voltage magnitudes are generated according to a prior distribution $p(\s,\bm{P}, \Q^{in}, \V^{in})$, and b) all the quantities $\h$ measured in the system can then be computed by solving the power flow equations \eqref{acpf}, based on which the actual noisy measurements $\y$ follow the conditional probability distribution $p(\y|\s, \bm{P}, \Q^{in}, \V^{in})$. We note that, as in many Bayesian inference problems, an accurate prior distribution $p(\s,\bm{P}, \Q^{in}, \V^{in})$ may be difficult to obtain in practice. Nonetheless, a sharp concentration of the posterior distribution on the true post-outage network topology allows effective inference of multi-line outages even in the absence of accurate knowledge of the prior. 
\end{RK}

Our objective is to infer the topology of the power grid $\s$ given the observed measurements $\y$. Thus, under a Bayesian inference framework, we are interested in computing the posterior \emph{conditional} probabilities: $\forall \s$,
\begin{align}
&p(\s|\y) \nn\\
&= \frac{\int p(\s,\bm{P}, \Q^{in}, \V^{in}) p(\y|\s, \bm{P}, \Q^{in}, \V^{in})d\P d\Q^{in}d\V^{in}}{p(\y)}. \label{apost}
\end{align}
Given the observations $\y$, a maximum a-posteriori probability (MAP) detector would pick $\argmax_{\bm{s}} p(\bm{s}|\bm{y})$ as the topology/multi-line outage identification decision, which minimizes the identification error probability \cite{poor1994introduction}. 
However, 
as the number of hypotheses of $\s$ grows exponentially with the number of unknown line statuses, performing such a hypothesis testing based on an exhaustive search becomes computationally intractable. 
\emph{In general, as there are up to $2^L$ possibilities for $\s$, computing, or even listing the probabilities $p(\s|\y), \forall \s$ has an exponential complexity.} 

\subsubsection*{Posterior Marginal Probabilites}
As an initial step towards addressing the fundamental challenge of computational complexity, instead of computing $p(\s|\y)$, we focus on computing the \emph{posterior marginal conditional probabilities} $p(s_l | \y), l = 1,\ldots,L$. We note that the posterior marginals are characterized by just $L$ numbers, $\PP(s_l = 1 | \y), l=1, \ldots, L$, as opposed to $2^L-1$ numbers required for characterizing $p(\s | \y)$. Accordingly, the hypothesis testing problem on $\s$ is decoupled into $L$ separate \emph{binary} hypothesis testing problems: for each line $l$, the MAP detector identifies $\argmax_{s_l\in\{0,1\}}p(s_l | \bm{y}, \bm{P})$. As a result, instead of minimizing the identification error probability of the vector $\bm{s}$, 
the binary MAP detectors minimize the identification error probability of each line status $s_l$. 

Although listing the posterior marginals $p(s_l | \bm{y})$ are tractable, \emph{computing} them, however, still remains intractable. In particular, even with $p(\s|\y)$ given, summing out all $s_k, k\ne l$, to obtain $p(s_l | \bm{y})$ still requires exponential computational complexity \cite{mezard2009information}. As a result, even a \emph{binary} MAP detection decision of $s_l$ cannot be made in a computationally tractable way. 
This challenge will be addressed by a novel method we will develop in the next section. 

\section{A Learning-To-Infer Method} \label{sec:lti}
\subsection{A Variational Inference-Inspired Framework}
In this section, we develop a variational inference-inspired method for approximate inference of the posterior marginal conditional probabilities $p(s_l | \y), l=1, \ldots, L$. The general idea is to find a 
conditional distribution $q(\s | \y)$ that 
\begin{itemize}
\item[a)] approximates the original $p(\s | \y)$ very closely, and 
\item[b)] offers fast and accurate multi-line outage identification results based on easily computable $q(s_l | \y),\forall l$. 
\end{itemize}
In particular, we consider that $q(\s | \y)$ is modeled by some parametric form (e.g., neural networks), and is hence chosen from some family of parametrized conditional probability distributions $\{q_\be(\s|\y)\}$, where $\be$ is a vector of model parameters. It is worth noting that $q(\s | \y)$ is a \emph{function of both $\s$ and $\y$}, and the parameters $\be$ associate both $\s$ and $\y$ with the probability value $q_\be(\s | \y)$, for all possible $\s$ and $\y$. 

To achieve the two goals above, we aim to choose a family of probability distributions $\{q_\be(\s|\y)\}$ to satisfy the following: 
\begin{itemize}
\item
The parametric form of $\{q_\be(\s|\y)\}$ has sufficient expressive power to represent very complicated functions, so that our approximation to the true $p(s_l | \y)$ can be made sufficiently precise.
\item
It is easy to compute the marginal $q_\be(s_l | \y)$, so that we can use it to infer $s_l$ with low computational complexity in real time based on the observed $\y$. 
\end{itemize}
From a family of parametrized distributions $\{q_\be(\s|\y)\}$, we would like to choose a $q_\be(\s|\y)$ that approximates $p(\s|\y)$ as closely as possible. For this, we employ the \emph{Kullback-Leibler (KL) divergence} as a metric of closeness between two probability distributions, 
\begin{align}\label{KLdef}
D(p\Vert q_\be) \triangleq \sum_{\s} p(\s|\y) \log \frac{p(\s|\y)}{q_\be(\s|\y)}. 
\end{align} 
Note that, for any particular realization of observations $\y$, a KL divergence $D(p\Vert q_\be)$ can be computed. Thus, $D(p\Vert q_\be)$ can be viewed as a function of $\y$. Since we would like the parametrized conditional $q_\be(\s|\y)$ to closely approximate $p(\s|\y)$ \emph{for all $\y$}, we would like to minimize the \emph{expected KL divergence} as follows: 
\begin{align}
& \min_\be ~\E_{\y}\left[ D(p\Vert q_\be) \right] \nn\\
\Leftrightarrow~ & \min_\be  ~ \sum_{\y} p(\y) \sum_{\s} p(\s|\y) \log \frac{p(\s|\y)}{q_\be(\s|\y)} \nn\\
\Leftrightarrow~ & \min_\be  ~ \sum_{\s,\y} p(\s,\y) \log \frac{p(\s|\y)}{q_\be(\s|\y)} \nn\\
\Leftrightarrow~ & \max_\be  ~ \E_{\s,\y}\left[  \log q_\be(\s|\y) \right], \label{KLopt}
\end{align}
where the expectation is taken with respect to the \emph{true} distribution $p(\s,\y)$. 

\begin{table}[t]
\normalsize
\caption{The Learning-to-Infer Method} \label{ltftable}
\centering
\begin{tabular}[c]{@{} l @{} l @{} p{4.0cm} @{}}
\hline
\multicolumn{3}{@{}l}{\emph{Offline computation:}}\\ 
~~~~~&1. Generate \emph{labeled} data set $\{\s^i,\y^i\}$ using Monte Carlo\\
&~~  simulations with the full-blown power flow and \\
&~~ sensor models. \\
~~~~~&2. Select a parametrized predictor model $\{q_\be(\s|\y)\}$. \\ 
~~~~~&3. Train the model parameters $\be$ using the generated data\\ 
&~~~set.\\
\multicolumn{3}{@{}l}{\emph{Online inference (in real time):}}\\ 
~~~~~&1. Collect instant measurements $\y$ from the system. \\
~~~~~&2. Compute the approximate posterior marginals \\
&~~~$q_{\be^*}(s_l|\y),l=1,\ldots,L,$ and infer the line statues $\{s_l\}$. \\ 
\hline
\end{tabular}
\end{table}

\subsection{From Generative Model to Discriminative Learning}
Evaluating $\E_{\s,\y}\left[  \log q_\be(\s|\y) \right]$ is, however, very difficult, primarily because it again requires the summation of an exponentially large number of terms. To address this, the key step forward is that we can \emph{approximate the expectation by the empirical mean} of $\log q_\be(\s|\y)$ over a large number of \emph{Monte Carlo samples}, generated according to (ideally) the \emph{true} joint probability $p(\s, \bm{P}, \Q^{in}, \V^{in}, \bm{y})$ (cf. \eqref{genmod}). We denote the relevant Monte Carlo samples by 
$\{\s^i, \y^i; i=1,\ldots, I\}$. 
Accordingly, \eqref{KLopt} is approximated by the following,  
\begin{align}
\max_\be  \frac{1}{I}\sum_{i=1}^I \log q_\be(\s^i|\y^i). \label{learn1}
\end{align}
With a data set $\{\s^i, \y^i\}$ generated using Monte Carlo simulations, \eqref{learn1} can then be 
solved as a deterministic optimization problem. 
The optimal solution of the model parameters $\be^*$ approaches that for the original problem \eqref{KLopt} as $I\rightarrow\infty$. 

In fact, the problem \eqref{learn1} can be viewed as an \emph{empirical risk minimization} problem in machine learning \cite{vapnik1998statistical}, as it trains 
a \emph{discriminative model} $q_\be(\s|\y)$ with a data set $\{\s^i, \y^i\}$ generated from a \emph{generative model} $p(\s, \bm{P}, \Q^{in}, \V^{in}, \y)$ (cf. Remark \ref{rkgen}). 
As a result of this offline learning / training process \eqref{learn1}, an approximate posterior function $q_{\be^*}(\s | \y)$ is obtained. Furthermore, it can be shown that \eqref{learn1} is equivalent to finding the maximum likelihood estimate of $\be$ on the data set $\{\s^i, \y^i\}$. 

\subsection{Offline Learning for Online Inference}
It is important to note that, 
\begin{itemize}
\item[a)] the training process to obtain the function $q_{\be^*}(\s | \y)$ is conducted completely \emph{offline};  
\item[b)] the \emph{use} of the trained function $q_{\be^*}(\s | \y)$ is, however, \emph{in real time, i.e., online}. 
\end{itemize}
In particular, in real time, given whatever \emph{newly observed measurements} $\y$ of the system, based on $q_{\be^*}(\s | \y)$, the \emph{approximate posterior marginals} $q_{\be^*}(s_l | \y), l = 1,\ldots,L$ will be computed. Based on such instantly computed $q_{\be^*}(s_l | \y)$, a detection decision of whether line $l$ ($=1,\ldots,L$) is connected or not in the current topology will be made. For example, a MAP detector would make the following decision, 
\begin{align}
\forall l = 1,\ldots,L, ~\hat{s}_l = 
\begin{cases}
0, & \mbox{ if } q_{\be^*}(s_l = 0 | \y) > 0.5, \\
1, & \mbox{ otherwise. } 
\end{cases}
\end{align}

\begin{figure}[t!]
		\centering
		\includegraphics[scale=0.57]{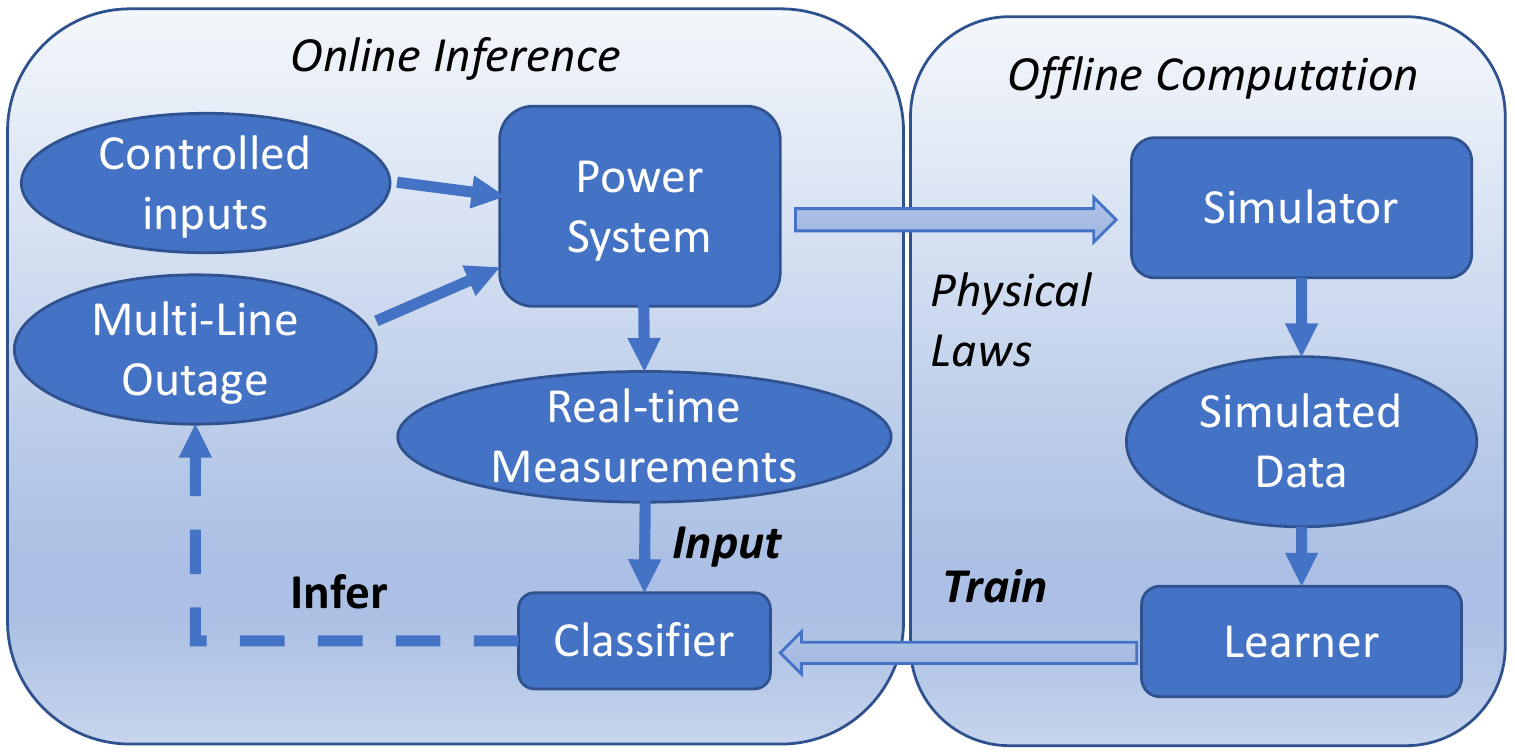} 
		\caption{Overall architecture of the Learning-to-Infer method.}
		\label{diagram}
\end{figure}

Accordingly, we name our proposed methodology ``\emph{Learning-to-Infer}'': To perform \emph{real time inference} of multi-line outages, we exploit \emph{offline learning} to train a detector based on \emph{labeled} data simulated from the full-blown physical model of the power system. The methodology is summarized in Table \ref{ltftable}. 
A system diagram is plotted in Figure \ref{diagram}. 

\begin{RK}[Training Binary Classifiers] \label{rkclass}
For any detector that identifies the status of a line $l$, (e.g., a binary MAP detector), it can also be viewed as a \emph{binary classifier} $\hat{s_l}(\y)\in \{0,1\}$: For each possible realization of $\bm{y}$, this classifier outputs an inferred status of line $l$. From this perspective, solving \eqref{learn1} is exactly a \emph{supervised learning} process based on a  \emph{labeled} data set, $\{\s^i, \y^i\}$, where $\{\s^i\}$ are the output labels that correspond to the input data $\{\y^i\}$. As a result, the rich literature on supervised learning for training binary classifiers directly apply to our problem under this Learning-to-Infer framework. 
\end{RK}

{
\begin{RK}[Difference from Variational Inference] It is worth noting the fundamental difference between the proposed Learning-to-Infer method and variational inference methods. Importantly, for every new inference instance given a new observation $\y$, variational inference methods need to call an optimization procedure to solve for a new variational model. In contrast, Learning-to-Infer only trains the predictor $q_\be(\s|\y)$ \emph{once} in an offline fashion, and simply calls the trained $q_\be(\s|\y)$ for any new inference instance given a new observation $\y$. As such, the online computation time needed by Learning-to-Infer is very little (e.g., performing a forward pass in a neural network), whereas that needed by variational inference methods is much more significant.  
In essence, Learning-to-Infer exploits the underlying lower dimensional structure of $p(\s|\y)$ to achieve \emph{generalizability} of the trained predictor $q_\be(\s|\y)$ to all possible new observations $\y$.  
\end{RK}
}

\subsection{Advantages of the Proposed Method} \label{sec:adv}
One great advantage of this Learning-to-Infer method is that we can generate \emph{labeled data} very efficiently. Specifically, we can efficiently sample from the generative model of $p(\s, \bm{P}, \Q^{in}, \V^{in}, \bm{y})$ (cf. \eqref{genmod}) as long as we have some prior $p(\s, \bm{P}, \Q^{in}, \V^{in})$ that is easy to sample from. While historical data and expert knowledge would surely help in forming such priors, using simple uninformative priors can already suffice as will be shown later in the numerical examples. 
As a result, we can obtain \emph{an arbitrarily large set of data at very little cost} to train the discriminative model. This is quite different from the typical situations encountered in machine learning problems, where obtaining a large amount of labeled data is usually expensive as it requires extensive human annotation effort. 

Furthermore, 
once the approximate posterior distribution $q_\be(\s|\y)$ is learned, it can be deployed to infer the multi-line outages  
\emph{in real-time} as the computation complexity of $q_\be(s_l |\y)$ is very low by design. This is especially important in monitoring large-scale power grids in real time, because, although training $q_\be(\s|\y)$ could take a reasonably long time, the inference speed is very fast. Therefore, the learned predictor $q$ can be used in real time with low-cost hardware. 

\subsubsection*{Limitations of Historical Data and Power of Simulated Data}
In overcoming the computational complexity challenges of real-time multi-line outage identification, it is particularly worth noting the fundamental limitation of using real historical data. Even with the explosion of data available from pervasive sensors in power systems, the data are often collected under a very limited set of system scenarios. For example, most historical data are collected under normal system topologies. Even with data collected under slowly updated systems or faulty systems, \emph{the underlying topologies in these real world cases only represent an extremely small fraction of the entire, exponentially large set of all topologies}.  
Consequently, historical data are fundamentally insufficient to resort to for real-time multi-line outage identification especially under rare failure events. 

Simulated data, as evidenced in the proposed Learning-to-Infer framework, offer great potential beyond what historical data can offer. An \emph{orders of magnitude richer set of scenarios} can be generated, and a learning procedure based on these simulated data can provide very powerful classifiers for identifying arbitrary multi-line outages that may appear in the future, but have not at all appeared in the past including the simulated scenarios. Last but not least, it is important to note that the simulated scenarios needed for the proposed Learning-to-Infer method would still be \emph{a very small fraction} of the entire, exponentially large model space, as will be demonstrated later in the numerical experiments. As such, it is the good \emph{generalizability} of the classifiers trained using the simulated data that enables effective outage inference under new failure events.

{
\begin{RK}[Learning from the Physical Model] In the proposed Learning-to-Infer method, the training process is at heart learning from the underlying power system physical model. Instead of manually deriving outage detection rules from analyzing the physical model, the proposed method uses a training procedure to learn such rules from massive data generated \emph{according to} the physical model. As such, the rich information embedded in the physical model are carried by the data simulated with it, and then learned by the predictor from training with these simulated data. The Learning-to-Infer method is thus a systematic ``indirect'' way of learning and using the information from the physical model. 
\end{RK}

\begin{RK}[Side Information and Change of Settings] 
An interesting question on generalizing the Learning-to-Infer method is how additional information (other than the observed $\y$) may be incorporated. For example, the system operator may receive the side information that certain lines are active for sure. Furthermore, there can also be more \emph{systematic} changes on what information are collected, notably, change of the measurement set $\y$ due to installation of additional sensors. 
For incorporating additional information, one way is to introduce additional inputs to the predictor during the offline training process. For example, we can let each line have a ``prior'' (even though in reality it can come from a posterior knowledge source) which is fed into the predictor. The data set generation and training would then need to include varying priors of these. Furthermore, a systematic way of dealing with slowly updating priors as well as changes in the measurement sets is to employ ``Transfer Learning''. Specifically, the changes in the measurement sets tend not to be so dramatic over a short period of time. Thus, the previously trained neural network can serve as a good initial point when we tune the neural network for an updated measurement set. The additional training time needed would be much shorter than if we train from scratch. These extensions are however out of the scope of this paper, and are left for future investigations. 
\end{RK}
}

\section{Neural Network Architectures for Learning Classifiers} \label{sec:NN}
To perform binary MAP inference of each line status, the decision boundary of the MAP detector is highly nonlinear (cf. Remark \ref{rkclass}). We investigate classifiers based on neural networks to capture such complex nonlinear decision boundaries. In other words, we employ neural networks as the parametric models $q_\be(\s|\y)$: given the input data $\y$, the output layer of the neural network will produce the probabilities $q_\be(s_l|\y), l=1,\ldots,L$, (based on which identification decisions are then made.) 

\begin{figure}[tb!]
	\centering
	\includegraphics[scale=0.23]{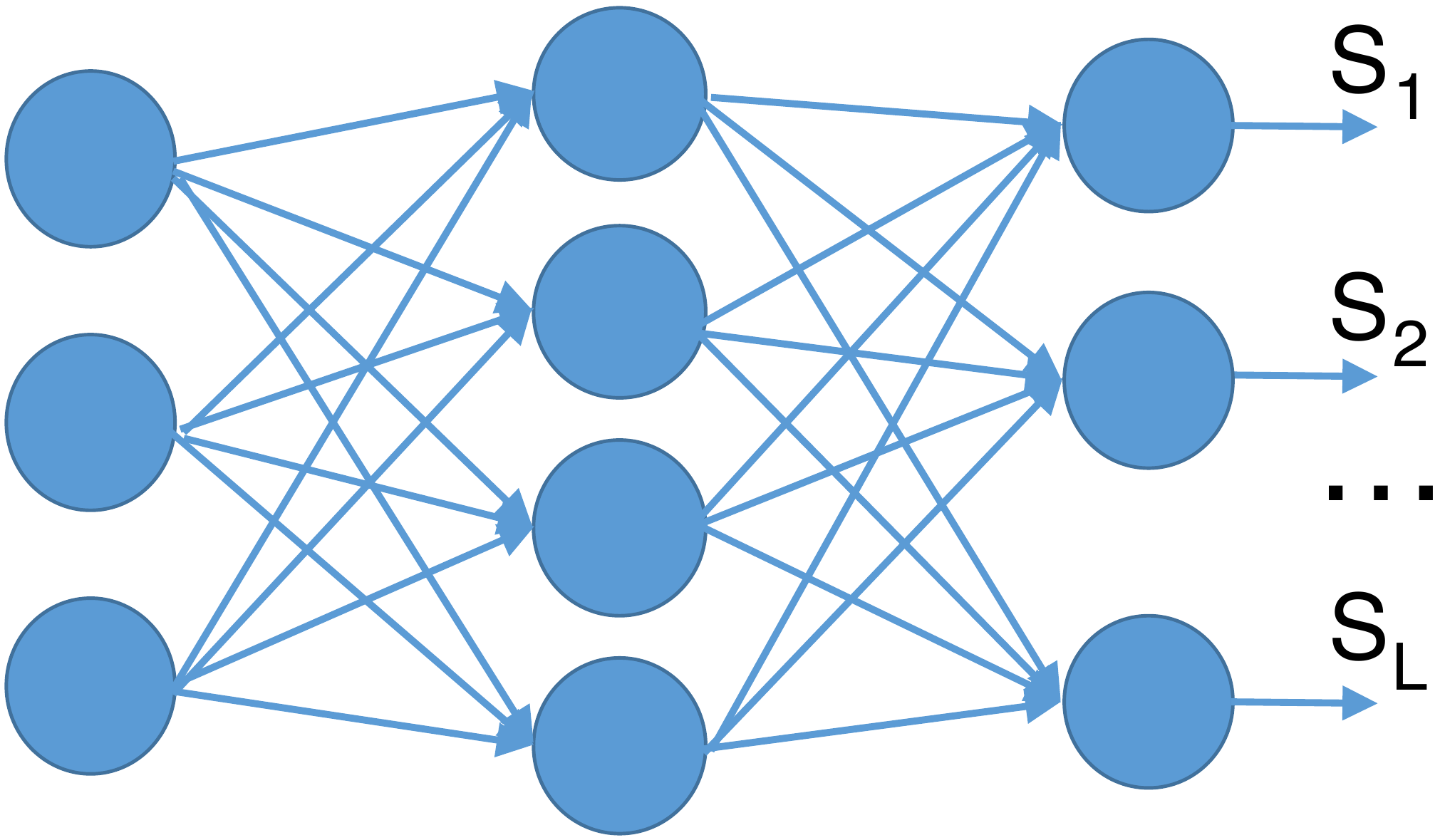}
	\caption{A single jointly trained neural network (which could have multiple hidden layers) whose features are shared for inferring all $L$ line statuses.}
	\label{sharedNN}
\end{figure}

In particular, we employ a neural network 
architecture that allows classifiers for different lines to \emph{share features}.  
Specifically, instead of training $L$ separate neural networks each with one node in its output layer, we train \emph{one} neural network whose output layer consists of $L$ nodes each predicting a different line's status. 
An illustration of this architecture is depicted in Figure \ref{sharedNN}: a) the input layer of the neural network consists of $\bm{y}$, b) 
the hidden layers of neurons compute a number of \emph{nonlinear features} of the input $\bm{y}$, and c) the output layer applies binary classifiers to these features to predict $s_l\in\{0,1\}, l=1,\ldots,L$. Specifically, logistic functions are employed in the output layer whose outputs correspond to $q_\be(s_l=1|\y), l=1,\ldots,L$. 
As a result, the features computed by the hidden layers can all be used in classifying any line's status. 
The intuition 
of using shared features is that certain \emph{common} features may provide good predictive power in inferring \emph{many different} lines' statuses in a power network. For training and testing, we generate labeled data $\{\bm{s}^i, \bm{y}^i\}$ randomly that satisfy the power flow equations and the observation models. Each $\bm{s}^i = [s^i_1, \ldots, s^i_L]^T$ then consists of $L$ labels used by the $L$ output classifiers respectively.

With the proposed Learning-to-Infer method, since labeled data can be generated in an arbitrarily large amount using Monte Carlo simulations, whenever overfitting is observed, it can in principle always be overcome by generating more labeled data for training. Thus, as long as the computation time allows, we can use neural network models of very high 
complexity for approximating the binary MAP detectors, without worrying about overfitting. 

\section{Numerical Experiments}\label{sec:num}
We evaluate the proposed Learning-to-Infer method for multi-line outage identification with three benchmark systems of increasing sizes, the IEEE 30, 118, and 300 bus systems, as the baseline topologies. 
As opposed to considering only a small number of 
simultaneous line outages as in existing works, we allow \emph{any number} of line outages, and investigate whether the learned discriminative classifiers can successfully recover the post-outage topologies in real time. 

\subsection{Data Set Generation} \label{sec:dgen}
In our experiments, the data sets are primarily generated with the DC power flow model \eqref{DC}. Here, our focus is to examine whether the proposed Learning-to-Infer method can effectively overcome the fundamental challenge of exponential computation complexity due to the potentially large number of simultaneous line outages. For this, the DC power flow model offers sufficient modeling details.  
We will then at the end of the section run experiments with data sets generated with the AC power flow model \eqref{acpf}, and verify that the lessons learned from the DC power flow experiments continue to hold. 

With the DC power model, 
the set of controlled inputs $\{\P, \Q^{in}, \V^{in}\}$ reduce to $\{\P\}$, and the generative model \eqref{genmod} reduces to $p(\s, \bm{P}, \bm{y}) = p(\s,\bm{P}) p(\y|\s, \bm{P})$. 
To generate a data set $\{\s^i,\P^i,\y^i, i = 1,\ldots,I\}$, we assume the prior distribution $p(\s,\P)$ factors as $p(\s)p(\P)$. As such, we generate the post-outage network topologies $\s$ and the power injections $\P$ independently: 
\begin{itemize}
\item We generate the line statuses $\{s_l\}$ using independent and identically distributed (IID) Bernoulli random variables, so that the average numbers of line outages are $7.8, 13.4$ and $11.6$ for the IEEE 30, 118 and 300 bus systems, respectively. These numbers of simultaneous line outages are significantly higher than those typically assumed in sparse line outage studies. 
We do not consider disconnected networks in this study, and exclude the line status samples if they lead to disconnected networks. 
As such, considering that some lines must always be connected to ensure network connectivity, after some network reduction, the equivalent networks for the IEEE 30, 118, and 300 bus systems have $38, 170$, and $322$ lines that can possibly be in outage, respectively. 
\item We would like our predictor to be able to identify multi-line outages for \emph{arbitrary values of power injections} as opposed to fixed ones. Accordingly, we generate $\P$ using the following procedure: For each data sample, we first generate bus voltage phase angles $\thet$ as IID uniformly distributed random variables in $[0, 0.2\pi]$, and then compute $\P$ according to \eqref{DC} under the baseline topologies. We note that, the 
spread of the phase angles in the generated data sets can 
cover nearly all possible power injection cases in real power transmission networks. 
\end{itemize} 
With each pair of generated $\s^i$ and $\P^i$, we consider two types of measurements that constitute $\y$: nodal voltage phase angle measurements and nodal power injection measurements. For these, a) we generate IID Gaussian voltage phase angle measurement noises with a standard deviation of $0.01$ degree, the state-of-the-art PMU accuracy \cite{von2013micro}, and b) we assume power injections are measured accurately. In the following experiments, we consider that measurements of voltage phase angles and power injections are collected at all the buses. The effect of number and locations of sensors will be discussed later in 
this section. 

\begin{table}[t]
\normalsize
\caption{Data set size vs. the entire search space} \label{sizetable}
\centering
\begin{tabular}[c]{| l | c | p{3.3cm} @{}}
\hline
\multicolumn{3}{|c|}{\emph{The (reduced) IEEE 30 bus system with $38$ lines}}\\ 
\hline
Number of all possible  & \multirow{2}{*}{$2^{38} = 2.75\times 10^{11}$}\\
post-outage topologies & \\
\hline
Number of topologies with ~~ & \multirow{2}{*}{${38 \choose 8} = 4.89 \times 10^7$}\\
$8$ line outages & \\
\hline
The generated data set & $3\times 10^5$ \\
\hline
\multicolumn{3}{@{}l}{}\\ 
\hline
\multicolumn{3}{|c|}{\emph{The (reduced) IEEE 118 bus system with $170$ lines}}\\ 
\hline
Number of all possible  & \multirow{2}{*}{$2^{170} = 1.50 \times 10^{51}$}\\
post-outage topologies & \\
\hline
Number of topologies with  & \multirow{2}{*}{${170 \choose 13} = 9.94 \times 10^{18}$}\\
$13$ disconnected lines &\\
\hline
The generated data set & $8\times 10^5$ \\
\hline
\multicolumn{3}{@{}l}{}\\ 
\hline
\multicolumn{3}{|c|}{\emph{The (reduced) IEEE 300 bus system with $322$ lines}}\\ 
\hline
Number of all possible & \multirow{2}{*}{$2^{322} = 8.54 \times 10^{96}$}\\
post-outage topologies & \\
\hline
Number of topologies with & \multirow{2}{*}{${322 \choose 12} = 2.11 \times 10^{21}$}\\
$12$ disconnected lines &\\
\hline
The generated data set & $2.2\times 10^6$ \\
\hline
\end{tabular}
\end{table}

\begin{figure*}[t!]
  \centerline{
    \subfigure[]{\includegraphics[scale=0.49]{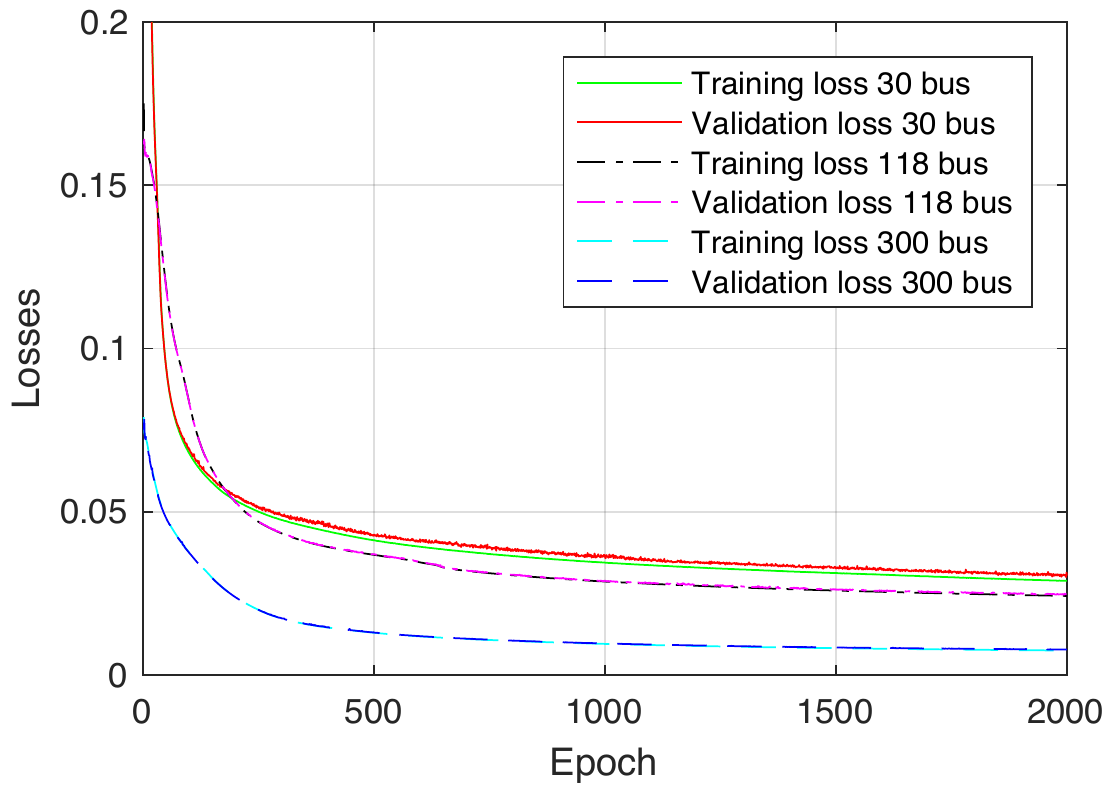} 
	\label{losscurves}
    } \subfigure[]{\includegraphics[scale=0.49]{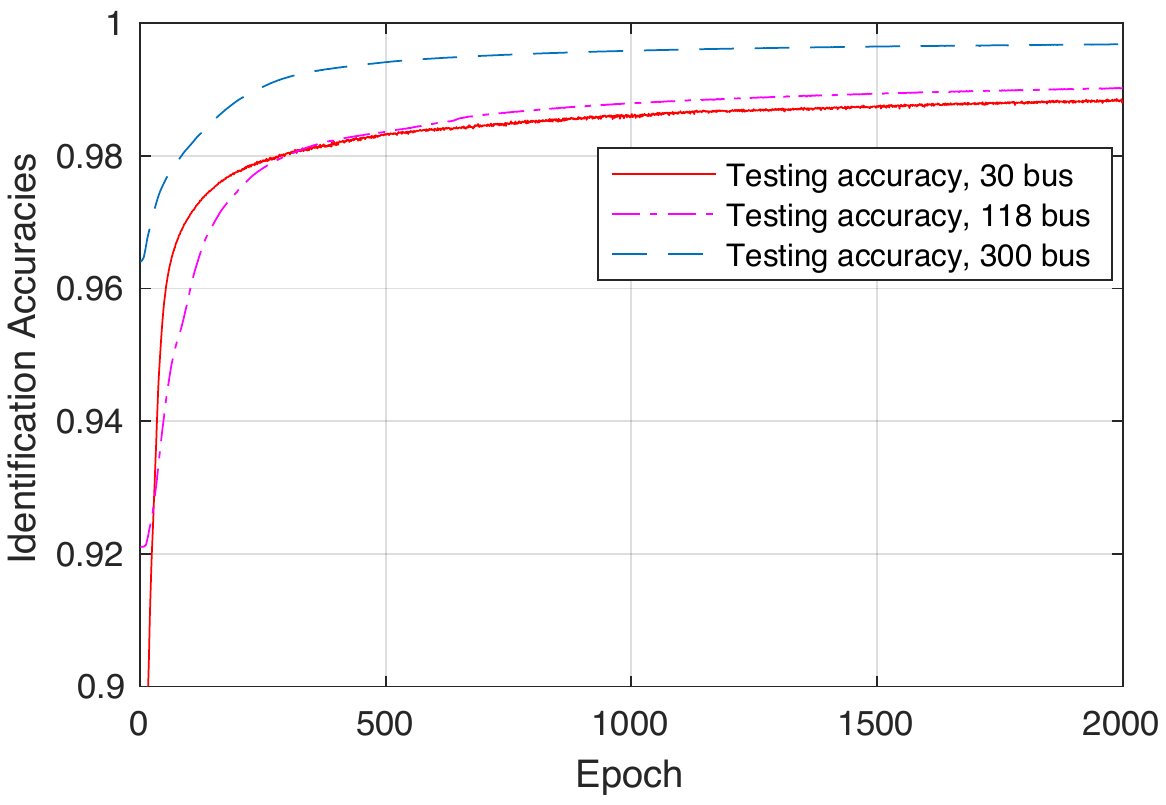} 
	\label{accucurves}
    } \subfigure[]{\includegraphics[scale=0.49]{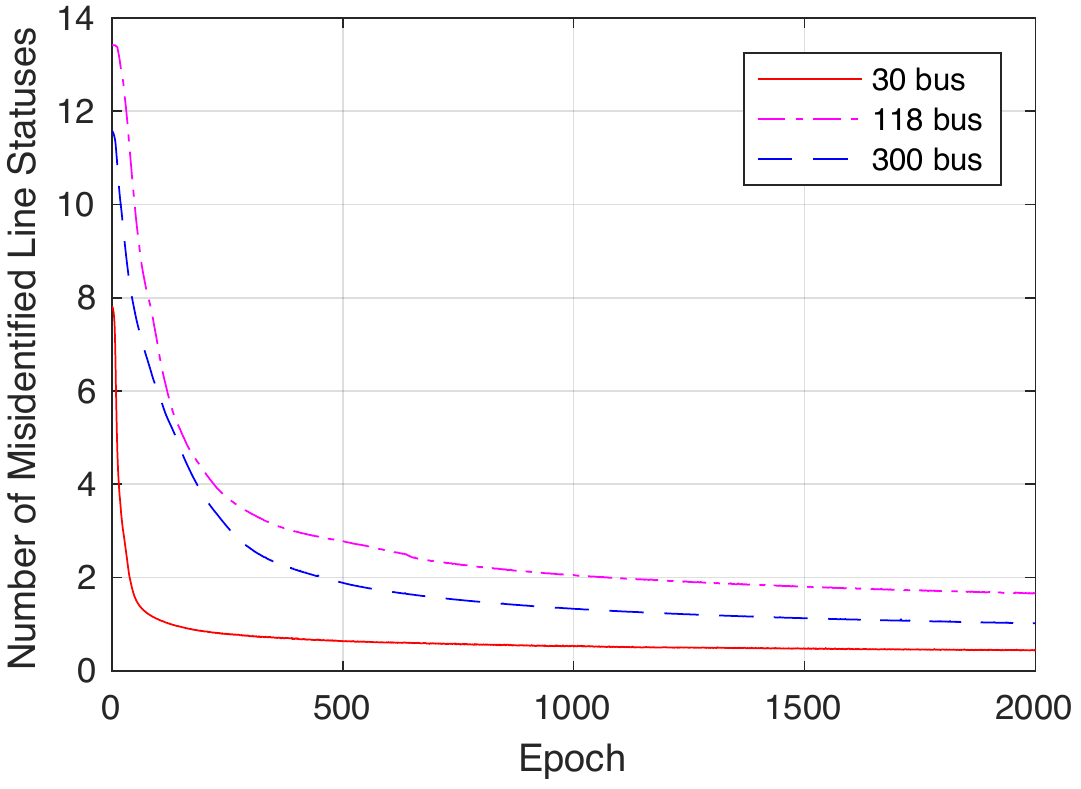} 
	\label{nlines}
    } }
  \caption{Progressions of a) training and validation losses, b) testing accuracies, and c) average numbers of misidentified line statuses in IEEE 30, 118 and 300 bus systems.}
\end{figure*}

In this study,  
we generate $300K$, $800K$, and $2.2M$ data samples for the IEEE 30, 118, and 300 bus systems, respectively. These $300K/800K/2.2M$ data 
are further divided into $200K/600K/1.8M$, $50K/100K/200K$, and $50K/100K/200K$ samples for training, validation, and testing, respectively. 
We note that \emph{over $99\%$} of the generated $300K$ 30-bus multi-line outages are distinct from each other, so are those of the generated $800K$ 118-bus multi-line outages and those of the $2.2M$ 300-bus multi-line outages. As a result, these generated data sets can very well evaluate the \emph{generalizability} of the trained classifiers, as (almost) all data samples in the test set have post-outage topologies \emph{unseen} in the training set.

Furthermore, we would like to compare the \emph{size of the generated data set} to the \emph{total number of possible outage hypotheses}, as highlighted in Table \ref{sizetable}. 
Clearly, a) it is computationally prohibitive to perform line outage inference based on exhaustive search, and b) \emph{the generated $300K, 800K$ and $2.2M$ data sets are only a tiny fraction of the entire space of all multi-line outages}. 
Yet, we will show that the classifiers trained with the generated data sets exhibit excellent inference performance and generalizability. 

\subsection{Neural Network Structure and Training}
We employ three-layer (i.e., one hidden layer) fully connected neural networks 
with the feature sharing architecture (cf. Figure \ref{sharedNN}). Rectified Linear Units (ReLUs) are employed as the activation functions in the hidden layer.  
In training the classifiers, we use stochastic gradient descent (SGD) with momentum update and Nesterov's acceleration \cite{nesterov2013introductory}. 
While this optimization algorithm works sufficiently well for our experiments, we note that other algorithms may further accelerate the training procedure \cite{kingma2014adam}.

\begin{figure*}[t]
  \centerline{
    \subfigure[]{\includegraphics[scale=0.48]{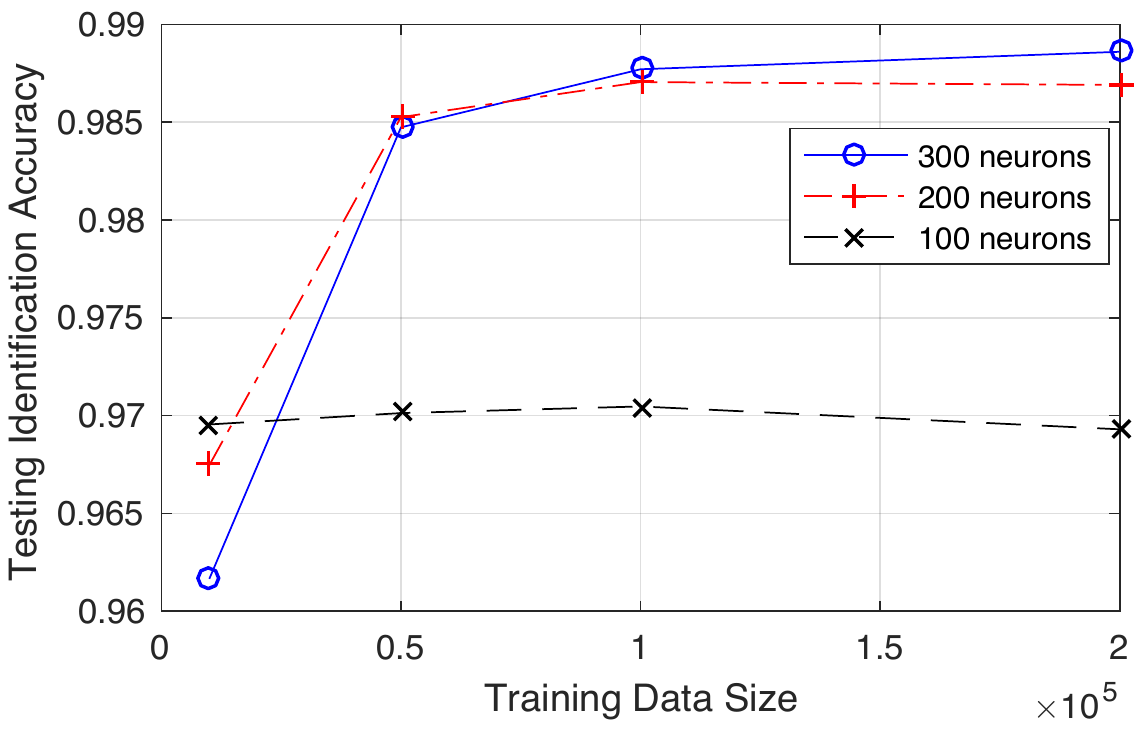} 
	\label{sampcomp30}
    } \subfigure[]{\includegraphics[scale=0.48]{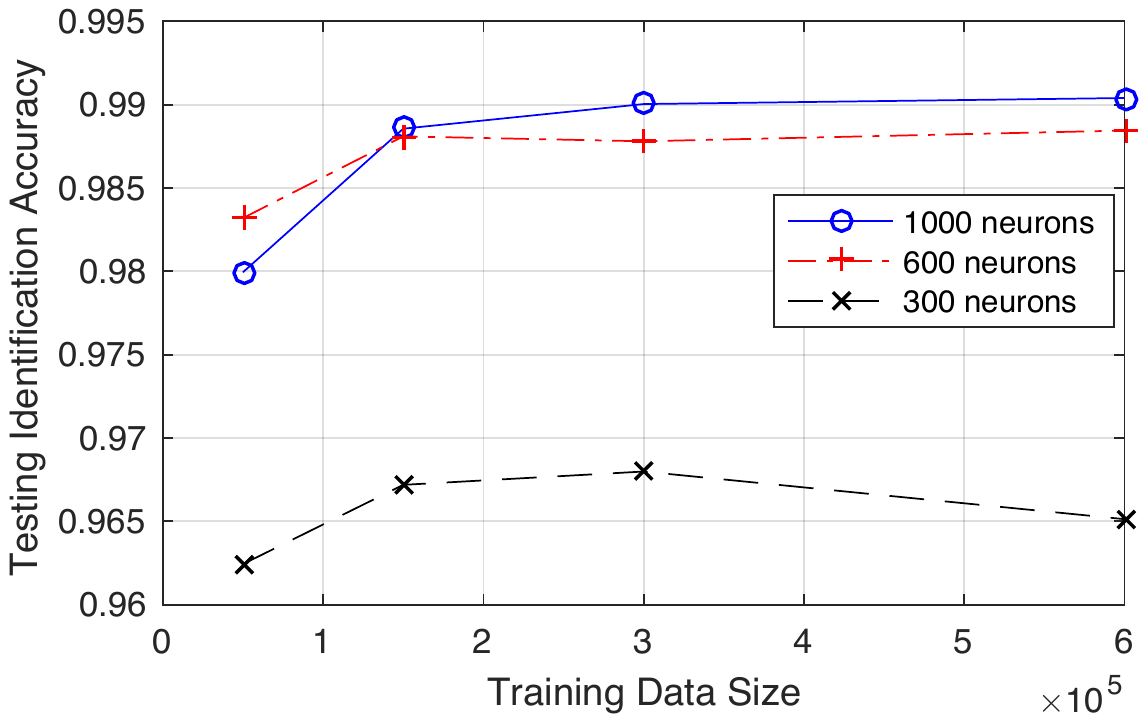} 
	\label{sampcomp118}
    } \subfigure[]{\includegraphics[scale=0.48]{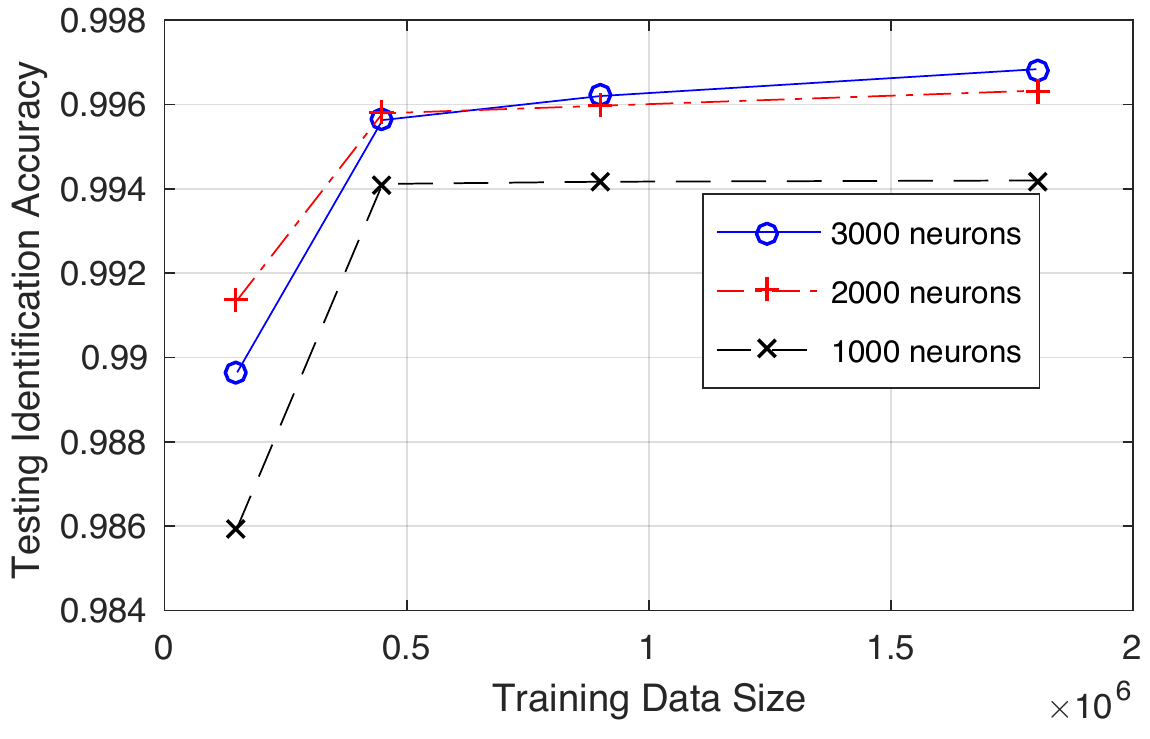} 
	\label{sampcomp300}
    } }
  \caption{Effect of model size and sample size, (a) IEEE 30 bus system, (b) IEEE 118 bus system, (c) IEEE 300 bus system.}
\end{figure*}

\subsection{Evaluation Results}

\subsubsection{Performance of the Learning-to-Infer Method} 

We employ $300, 1000$ and $3000$ neurons in the hidden layer for the IEEE 30, 118 and 300 bus systems, respectively. For all the three systems, we plot in Figure \ref{losscurves} the achieved training and validation losses for every epoch, and in Figure \ref{accucurves} the achieved testing accuracies for every epoch. It is clear that the training and validation losses stay very close to each other for all the three systems, and thus no overfitting is observed. Moreover, very high testing accuracies, 0.989, 0.990 and 0.997 are achieved for the IEEE 30, 118 and 300 bus systems, respectively. 

The testing accuracies can be equivalently understood by the \emph{average numbers of misidentified line statuses}, plotted in Figure \ref{nlines}. We observe that, \emph{at the beginning} of the training procedures, the average numbers of misidentified line statuses are $7.8$, $13.4$ and $11.6$ for the IEEE 30, 118 and 300 bus systems, which are exactly the \emph{average numbers of disconnected lines} in the respective generated data sets (cf. Section \ref{sec:dgen}). Indeed, this coincides with the result from a naive identification decision rule of always claiming all the lines as connected (i.e., a trivial majority guess). As the training procedures progress, the average numbers of misidentified line statuses are drastically reduced to eventually $0.4$, $1.7$ and $1.0$. In other words, for the IEEE 300 bus system for example, \emph{facing on average $11.6$ simultaneous line outages, only $1$ line status would be misidentified} on average by the learned classifier. We note that 
such a performance is achieved with outage identification decisions made \emph{in real time, under a millisecond}. 
While the training process can potentially be time consuming, it is however done completely offline. 

It is 
worth noting that we have generated the training, validation and testing data sets with uniformly random voltage phase angles, and hence considerably variable power injections. In practice, there is often more informative prior knowledge about the power injections based on historical data and load forecasts. With such information, the model can be trained with much less variable samples of power injections, and the outage identification performance can be further improved.

\subsubsection{Model Size, Sample Complexity, and Scalability}
In the proposed Learning-to-Infer method, obtaining labeled data is not an issue since data can be generated in an arbitrarily large amount using Monte Carlo simulations. This leads to two questions that are of particular interest: to learn a good classifier, a) what size of a neural network is needed? and b) how much data need to be generated? To answer these questions, 
we vary the sizes of the hidden layer of the neural networks as well as the training data size, and evaluate the learned classifiers for the three benchmark systems. 
We plot the testing results for the IEEE 30, 118 and 300 bus systems in Figure \ref{sampcomp30}, \ref{sampcomp118} and \ref{sampcomp300}, respectively. 
It is observed that the best performance is achieved with $200K/600K/1.8M$ data and with 300/1000/3000 neurons for the 30/118/300 bus systems, respectively. 
Further increasing the data size or the neural network size would see much diminished returns.

\begin{figure}[t!]
	\centering
	\includegraphics[scale=0.45]{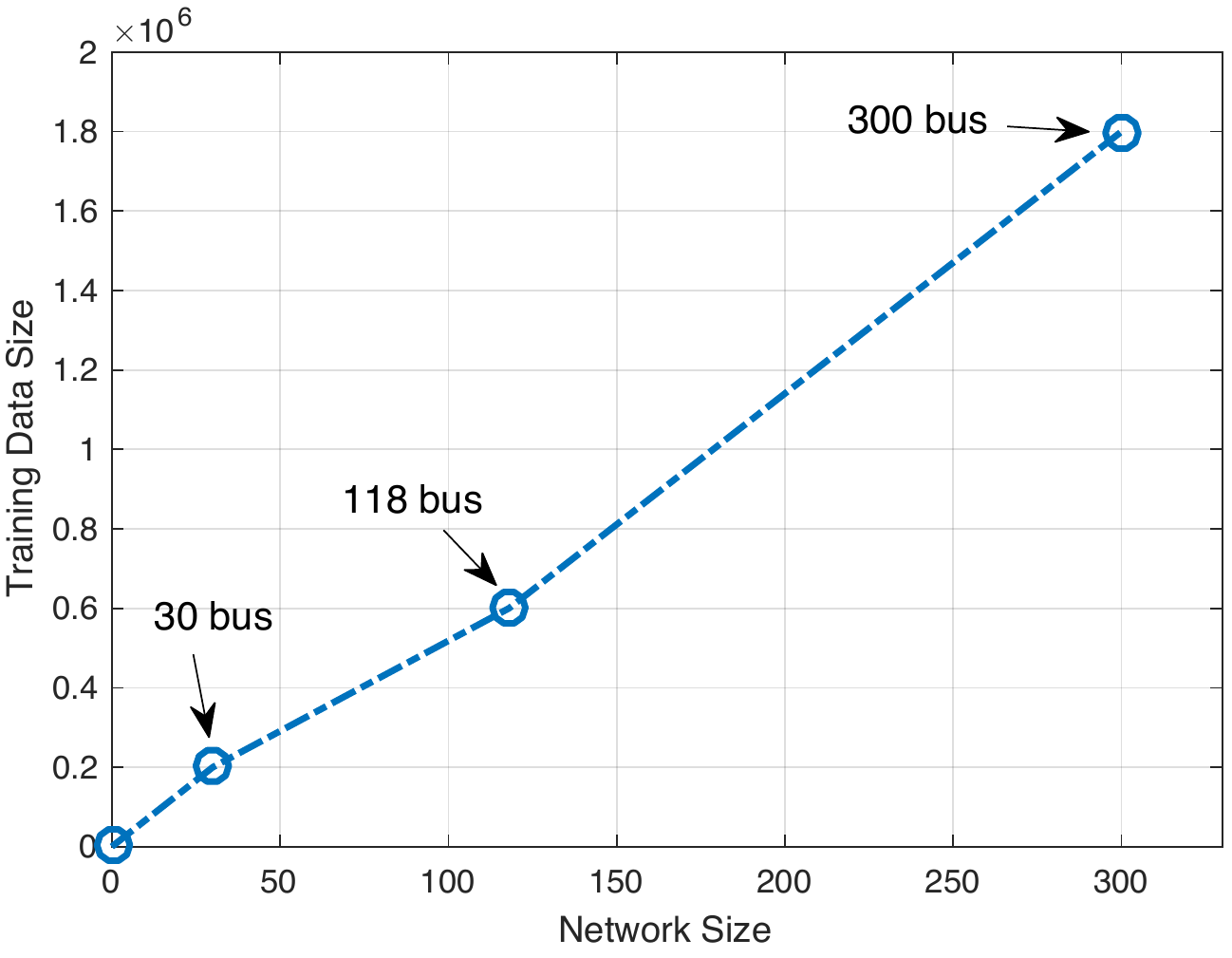} 
	\caption{Scalability of the Learning-to-Infer method, from the IEEE 30 bus system to the IEEE 300 bus system.}
	\label{scaling}
\end{figure}

Based on all these experiments, we now examine the \emph{scalability} of the proposed Learning-to-Infer method as the problem size increases. We observe that training data sizes of $200K, 600K$ and $1.8M$ and neural network models of sizes 300, 1000 and 3000 ensure very high and comparable performance with no overfitting for the IEEE 30, 118 and 300 bus systems, respectively. When these data sizes are reduced by a half, some levels of overfitting then appeared for these models in all the three systems. We plot the training data sizes compared to the problem sizes for the three systems in Figure \ref{scaling}. We observe that the required training data size increases approximately \emph{linearly} with the problem size. This linear scaling behavior implies that the proposed Learning-to-Infer method can be effectively implemented for large-scale systems with reasonable computation resources. 

\begin{figure}[tb!]
	\centering
	\includegraphics[scale=0.40]{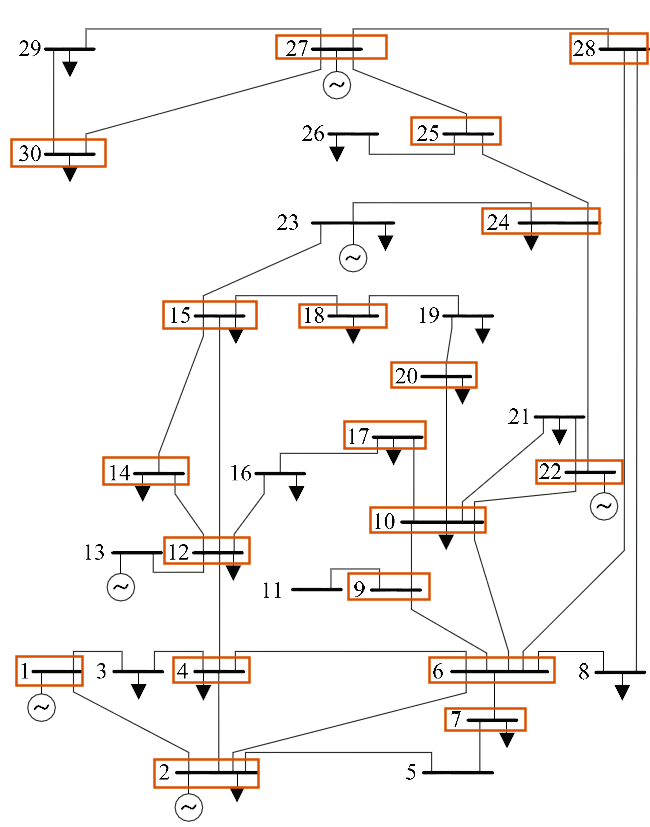}
	\caption{The IEEE 30 bus system, and a set of locations of PMUs.}
	\label{30bus}
\end{figure}

\subsubsection{Effect of Number and Locations of Sensors}
We now discuss 
the effect of sensor placement in real-time multi-line outage identification. It is clear that the performance of line outage identification would closely depend on where and what types of sensor measurements are collected. Given limited sensing resources, optimizing the sensor placement is a hard problem for which many studies have addressed (see, e.g., \cite{JSTSP14} among others). Here, we present the results from a case study on the IEEE 30 bus system, for which voltage phase angles are collected only at $19$ buses (as opposed to all the buses as in the previous experiments), as depicted in Figure \ref{30bus}. 
Interestingly, the achieved average identification accuracy only drops to $0.978$ (from $0.989$ when all the buses are monitored.) This translates to on average only $0.83$ misidentified line statuses among a total of $38$ lines. A more comprehensive study of sensor placement for real-time multi-line outage identification is left for future work. 

\begin{figure}[t!]
	\centering
	\includegraphics[scale=0.6]{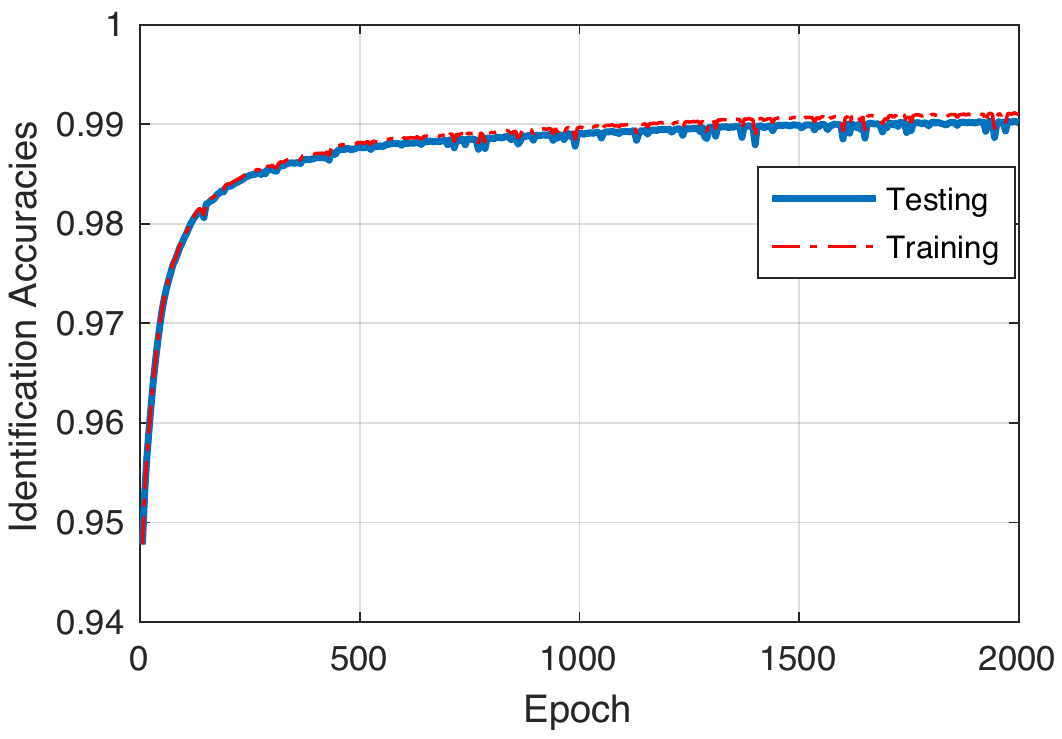} 
	\caption{Progressions of the training and testing accuracies, the IEEE 118-bus system, with the AC power flow model employed.}
	\label{ACcurves}
\end{figure}

\subsection{Experiments with the AC Power Flow Model} \label{sec:ac}
We close this section by verifying the performance of the proposed Learning-to-Infer method with data generated from the AC power flow model. Specifically, we consider the IEEE 118-bus system with 18 generators and 99 loads. Similar to the earlier data set generation process with the DC power flow model, we randomly generate $1M$ distinct connected post-outage topologies with an average number of $16.2$ line outages. We then significantly and randomly vary the power generation and loads in the system with standard deviations equal to 50\% of the means, and generate $1M$ distinct generation and load profiles.  

For each data point which includes a post-outage topology and a generation and load profile, we solve the AC power flow equations \eqref{acpf}. To have a consistent comparison with the earlier experiments with the DC power flow model, we continue to rely on measurements of nodal voltage phase angles, real power generation, and real power loads to infer the multi-line outages in real time. We will demonstrate that, with the AC power flow model, very high performance similar to that with the DC power flow model can be achieved. Undoubtedly, other types of measurements (e.g., voltage magnitudes, reactive power) may be used to further improve the performance, which is left for future investigation. 

The $1M$ data are divided into $800K$, $100K$, and $100K$ for training, validation, and testing, respectively. Similarly to the DC power flow experiments, we employ a two-layer fully connected neural network with 1000 neurons in the hidden layer for learning to infer multi-line outages. The same training algorithm is applied. We plot the training and testing accuracies for every epoch in Figure \ref{ACcurves}. We observe that a $0.990$ testing accuracy is achieved, (recall that the same accuracy, $0.990$, is achieved in the earlier experiments on the 118-bus system with the DC power flow model). This translates to on average $1.74$ mis-identified line statuses. 

Furthermore, we looked into the types of mis-identification errors, and observed that a) the rate of \emph{missed detection} (i.e., missing a line outage when it actually occurred among other simultaneous line outages) is $8.4\%$, and b) the rate of \emph{false alarm} (i.e., identifying a line as in outage when it is in fact connected) is a much lower $0.24\%$. As a result, we observe that nearly $80\%$ of the on average $1.74$ mis-identified line statuses are from \emph{missing to detect} $8.4\%$ of the on average $16.2$ simultaneous line outages, resulting in $1.36 (= 16.2 \times 8.4\%)$ missed line outages. 

{
\subsection{On Computation Times for Data Generation and Training}
As discussed above, a major advantage of the Learning-to-Infer method is that offline computation is exploited for achieving fast and accurate online inference. Specifically, the offline computation consists of two components: a) data generation based on the physical model, and b) predictor training based on the generated data. 
We discuss in the following several aspects of the offline computation times for data generation and predictor training. 

The time consumed for generating the 1M data with the AC power flow on the IEEE 118 bus system (cf. Section \ref{sec:ac}) is a little over an hour using MATPOWER \cite{MatP11}. The training time with 2000 epochs on these data is a little over two hours. Both are run on a laptop with an Intel Core i7 3.1-GHz CPU and 8 GB of RAM. Various approaches can be applied to reduce both times. On the one hand, data generation can be \emph{trivially parallelized} and significantly accelerated as such. It is worth re-emphasizing that data generation via simulations, while still may take a non-trivial amount of time for large systems, is regardless many orders of magnitude faster than collecting and manually labeling historical data from real-world systems. On the other hand, the experiments conducted in this section have achieved very high identification accuracies around or above $99\%$. In practice, if the performance requirement is not as high (e.g., $97\%$), then a significantly smaller amount of data (cf. Figures \ref{sampcomp30} \ref{sampcomp118} and \ref{sampcomp300}) and less number of training epochs (cf. Figure \ref{accucurves}) would be sufficient. The sizes of the neural networks can also be reduced which will lead to faster training. Leveraging the above approaches, much less computation times can be achieved for offline data generation and training. 
}

\section{Conclusion} \label{sec:concl}
We have developed a new Learning-to-Infer method for real-time multi-line outage identification in power grids. The computational complexity due to the exponentially large number of outage hypotheses is overcome by efficient marginal inference with optimized predictor models. Optimization of the predictor model is transformed to and solved as a discriminative learning problem, based on Monte Carlo samples efficiently generated with full-blown power flow models. The developed Learning-to-Infer method has the major advantages that a) the training process takes place completely offline, and b) labeled data sets can be generated in an arbitrarily large amount fast and at very little cost. As a result, very complex predictor models can employed without worrying about overfitting, as more labeled training data can always be generated had there been overfitting observed. With the classifiers learned offline, their actual use is in real time, and outage identification decisions are made under a millisecond. 
We have evaluated the proposed method with the IEEE 30, 118 and 300 bus systems. It has been demonstrated that arbitrary multi-line outages can be identified in real time with excellent performance using classifiers trained with a reasonably small amount of generated data. 

\ifCLASSOPTIONcaptionsoff
\newpage
\fi

\bibliographystyle{IEEEtran}
{\bibliography{Paper}}

\end{document}